# Employee Turnover Analysis Using Machine Learning Algorithms

Mahyar Karimi
Computer Engineering and Information Technology
Amirkabir University of Technology (Tehran Polytechnic)
Tehran, Iran
mahyarkarimi@aut.ac.ir

Kamyar Seyedkazem Viliyani
Faculty of Engineering
University of Tehran
Tehran, Iran
kamyarviliyani@ut.ac.ir

***Employee's knowledge is an organization asset. Turnover may impose apparent and hidden costs and irreparable damages. To overcome and mitigate this risk, employee's condition should be monitored. Due to high complexity of analyzing well-being features, employee's turnover predicting can be delegated to machine learning techniques. In this paper, we discuss employee's attrition rate. Three different supervised learning algorithms comprising AdaBoost, SVM and RandomForest are used to benchmark employee attrition accuracy. Attained models can help out at establishing predictive analytics.***

***Keywords— Employee Turnover; Supervised Learning; AdaBoost; Support-Vector Machines (SVM); RandomForest;***

## I. INTRODUCTION

Forty years ago, people feared that technology would reduce the need for educated workers, leaving large segments of the population unemployed. The opposite occurred. The demand for educated workers exceeds the supply. In the knowledge economy, companies are competing in two markets, one for its products and services and one for the talent required to develop and deliver them. With current low unemployment, the talent market is all the more competitive [1].

Recruitment and retention are now as important as production and distribution in the corporate business strategies of knowledge-intense companies [1]. One of the important assets of enterprises and organizations are people. Although most companies understand the importance of attracting and retaining talent, many lack a coherent approach to achieve their talent goals. Further, most lack a vision of how to integrate a system of practices to achieve their workforce objectives [1].

If employees are not satisfied, they leave the company or make others leave the company, so different and frequent rotations may reduce efficiency [2]. In spite of existence and development in information systems and knowledge bases, dependency on employees is not exterminated. This has its roots in integration of individuals with visions and characteristics of company. Education, talents, personality trains and etc. are criteria which an individual relies on. This issue could become a costly cycle and damage both the financial and spiritual aspects of the organization [2].

Imposed Costs when employees leave are listed below:

- Recruitment costs of new employees
- Other employees efficiency and effectiveness reduction cost
- Efficiency and effectiveness reduction of other employees
- Learning and training cost for the new employee
- The cost of relations and information loss
- The cost of abusing competitors
- The motivation cost for remaining employees

In this paper, we have applied comparative analysis using machine learning algorithms to create predictive model on employee's turnover. Dataset used in this study contains 30 features. Original dataset included 5 more features which was removed due to its inconvenient relation and lack of generalization. In the following, explanation and preprocessing on dataset is discussed. We have chosen three algorithms with the approach of maximizing accuracy on training data considering avoidance on overfitting. Two out of three, comprises AdaBoost and RandomForest from ensemble learning methods, and SVM.

## II. EFFECTIVE CRITERIAS ON TURNOVER

The first step in reducing turnover [3] is to find out why employees are leaving [4].

Turnover rates fluctuate from year to year, about 1.4% of an organization's employees leave each month (16.8% per year). Though some sources estimate that the cost of turnover can exceed 100% of the employee's annual salary [4][5][6], a review of 30 case studies suggests that the actual cost is probably closer to 20% of the annual salary for the position [7]. Both visible and hidden costs determine this estimate. Visible costs of turnover include advertising charges, employment agency fees, referral bonuses, recruitment travel costs, salaries and benefits associated with the employee time spent processing applications and interviewing candidates, and relocation expenses for the new employee. Hidden costs include the loss of productivity associated with the employee leaving—other employees trying to do extra work, no productivity occurring from the vacant position—and the lower productivity associated with a new employee being trained [4].

The relationship between job satisfaction and performance is not consistent across people or jobs. For example, for complex jobs, there is a stronger relationship between job satisfaction and performance than for jobs of low or medium complexity [4].

Many studies have shown that employee turnover significantly affects an organization's performance [8].

Measuring employee satisfaction is something companies have been doing for a long time, mainly using employee surveys and benchmarking information [9].

An annual employee survey is not agile or granular enough to give most businesses the timely, detailed insights they need to monitor and improve employee engagement. After all, employee opinions shift constantly, and change can happen very rapidly in many organizations. There is also the issue that most people hate filling out these lengthy surveys or, worse, they are worried their answers could be traced back to them, so they simply say what they think the company wants to hear. It is, therefore, questionable how useful the feedback even is [9].

There are a variety of research studies correlating employee engagement levels with things like innovation, safety, quality, revenue, shareholder value, customer satisfaction, retention, and so on. Those are the metrics we need to be talking about when we discuss engagement, especially if we want leaders to buy in and support this idea [10].

An opportunity for AI to support this would be to pull that disparate measures into a dashboard and show as close to real-time as posable which was correlated in a particular business and to what extent [10].

Across the spectrum of human capital management, from talent acquisition and employee development to talent mobility and engagement, greater demands are being placed on the human resources function to deliver tangible, actionable business results [10].

Employee satisfaction and retention are intrinsically linked. Gathered data from over 500,000 employees show that the attrition rate of employees with low engagement or satisfaction scores are 12 times higher than those with positive engagement scores. We also know that losing employee is costly (not to mention disruptive and time-consuming) [9].

## III. Concepts of Factors and Predictors

Concepts of different aspect of factors are discussed in three sections.

### *A. Concept 1*

Identifying predictors and causal factors in this instance, we are trying to find out what variables are linked to each other. Data can be tied together by correlation or causation. Correlation means that there seems to be a relationship in the data (for example, people seem to carry umbrellas more often when Rain is predicted). Causation is something else entirely (carrying an umbrella is not an enough reason to make weather rain more frequently). If we can identify what variables feed into others, then we can use those drivers or levers to create the results we need. For instance, if we can link increased training to higher sales, then that would seem to be a causal factor. Testing would need to be done to determine the extent of the linkage, but you get the picture. The first, most basic step for prediction is funding those predictors. Then we build on top of that foundation [10].

### *B. Concept 2*

Predictive modeling this takes the conversation a step further. Let me clarify quickly – We are talking here about leading variables. When a leading variable changes, it affects other elements down the line. Think about it as an assembly line – if we change something on the front end, the rest of the process is affected dramatically. If we mess up on the front end, then the rest of the process is affected differently. That is how we use leading variables within the predictive conversation. Once we have identified the predictor variables from the previous section, we step up a notch and start trying to predict what happens if we change one of our predictor variables. For instance, if we continue with the training/sales link mentioned above, the goal might be to try and see if doubling training also doubles sales. Alternatively, in another context, maybe we find out that there is a link between manager communication and employee engagement. Then we start trying to model whether increasing or decreasing manager communications affects engagement and to what extent. The point here is to focus on the driver variables and determine what happens if we start changing them around. How do they impact the final result? What changes occur [10]?

### *C. Concept 3*

Predicting behaviors: The final and most complex piece is determining what happens if we apply our model to new data or populations. In other words, can we predict how people will respond to certain variables? Let us say we have data on employee turnover that is related to a variety of factors, including manager check-ins, performance evaluation scores, and tenure. By mapping all of those variables for existing employees, we can create a model that will allow us to predict future behaviors. For instance, if the data shows that fewer manager check-ins, shorter tenure, and lower performance scores indicate someone is more likely to leave, we can put that person in a 'high risk' bucket. The person is more likely to leave than someone that does not have those types of factors working against them [10].

## IV. Introduction on Learning Models

Classifiers based on decision trees are attractive because of their straight forward and appealing idea. The classification in models which is created by decision trees are fast enough to provide answers for our need [11]. Studies on developing methods and evolvement of new techniques based on tree construction, its relationship to other algorithms like HMM [12] and multi-layer perceptrons, have took place for three decades. Reaching optimal accuracy on model and construction of a tree with minimal size are two subjects which many studies have focused on [11]. However, there is inclination on being adapted to the training data. This causes existence of unseen conditions due to lack of complete data. Pruning application on fully grown tree, in spite of reduction on training accuracy, results in growth of generalization accuracy. Although probabilistic methods does not guarantee optimization of the training set accuracy [11]. Trees should not grow too complex to avoid overfitting on training data. To overcome problems mentioned earlier, oblique decision trees are proposed. Using oblique hyperplanes usually outcomes in construction of a smaller tree. This results in fully splitting the data to leaves containing a single class [11].

The second algorithm used in this paper is based on concept of the wisdom of crowd. AdaBoost is a form ensemble method derived for Boosting algorithms [13]. By using weak learning algorithms which performs slightly better than random predicting, the model will reach expected accuracy on training data [13]. Weak models produced in each iteration are boosted with associated weights. Boosting refers to this general problem of producing a very accurate prediction rule by combining rough and moderately inaccurate rules-of-thumbs [13]. In each iteration, loss of accuracy caused by incorrectly classified data are adjusted by assigning theta values indicating error rate. Assigned thetas to each data of training set are updated through each iteration based on correctness of each data classification.

Boosting algorithms reduce bias and variance in a supervised learning. Boosting algorithms strongly avoid overfitting on training data, even though model continues to increase number of its classifiers.

Support vector machines which was proposed by Vapnik in 1995 implements the idea of mapping the input vectors into some higher dimensional feature space with non-linear mapping [14]. SVM not only solves the problem of finding a separating hyperplane, but also it will find the optimal separator [14]. For a two dimensional space, an optimal hyperplane with maximum distance between vectors is defined as a linear decision function separating training data space into two classes. Such optimal hyperplanes are yielded by using only a small subset of training data called support vectors [14]. Support vectors are marginal data. The expectation value of probability of committing an error on a test instance is bounded by the ratio between the expectation value of the number of support vectors and the number of trading vectors if the optimal hyperplane separates training data without error [14].

$$E\big(P(err)\big) = \frac{E[\#SupportVectors]}{E[\#TrainingVectors]} \quad (1)$$

It was shown that SVM treats computation problem of high dimensional spaces [15]. Support Vectors of a linearly separable training data can be calculated with linear kernel. To achieve specific value of classification error tolerance, linear kernel extended with hinge function. Non-linear classification is efficiently performed using what is called the kernel trick, which implies mapping input vectors into another space with higher dimension.

Radial Bias Function (RBF) kernel is used in SVM learning algorithm which maps the input into infinite dimensional space.

$$K(x,y) = e^{\left(\frac{\|x-y\|^2}{2\sigma^2}\right)} \quad (2)$$

## V. DATASET DESCRIPTION

Original dataset used in this study, which was introduced by IBM, was at first had 35 features. Five irrelevant features removed to reduce computation complexity and avoid model overfitting.

At first “EmployeeNumber” removed as it is a unique id on instances. “EmployeeCount” is removed because for all of the instances their value is 1. Due to same scenario for “Over18” and “StandardHours” feature, it is also removed. To generalize predictive model and accuracy on training data, we have removed “Department”.

LIST OF FEATURES OF DATASET

| | |
|---|---|
| Attrition | Age |
| BusinessTravel | DailyRate |
| DistanceFromHome | Education |
| EducationField | EnvironmentSatisfaction |
| Gender | HourlyRate |
| JobInvolvement | JobLevel |
| JobRole | JobSatisfaction |
| MaritalStatus | MonthlyIncome |
| MonthlyRate | NumCompaniesWorked |
| OverTime | PercentSalaryHike |
| PerformanceRating | RelationshipSatisfaction |
| StockOptionsLevel | TotolWorkingYears |
| TrainingTimesLastYear | WorkLifeBalance |
| YearsAtCompany | YearsInCurrentRole |
| YearsSinceLastPromotion | YearsWithCurrManager |

TABLE I. ALL FEATURES OF IBM DATASET EXCEPT THOSE DISCUSSED ABOVE.

Some of the features were labeled with string classes, they have been replaced with numerical value to prepare data to be ready for calculations. In addition, normalization on values of each feature applied to inhibit outliers. Following in this paper, correlation between features is shown in format of heatmap.

## VI. CORRELATION OF FEATURES

### A. *Pearson Correlation*

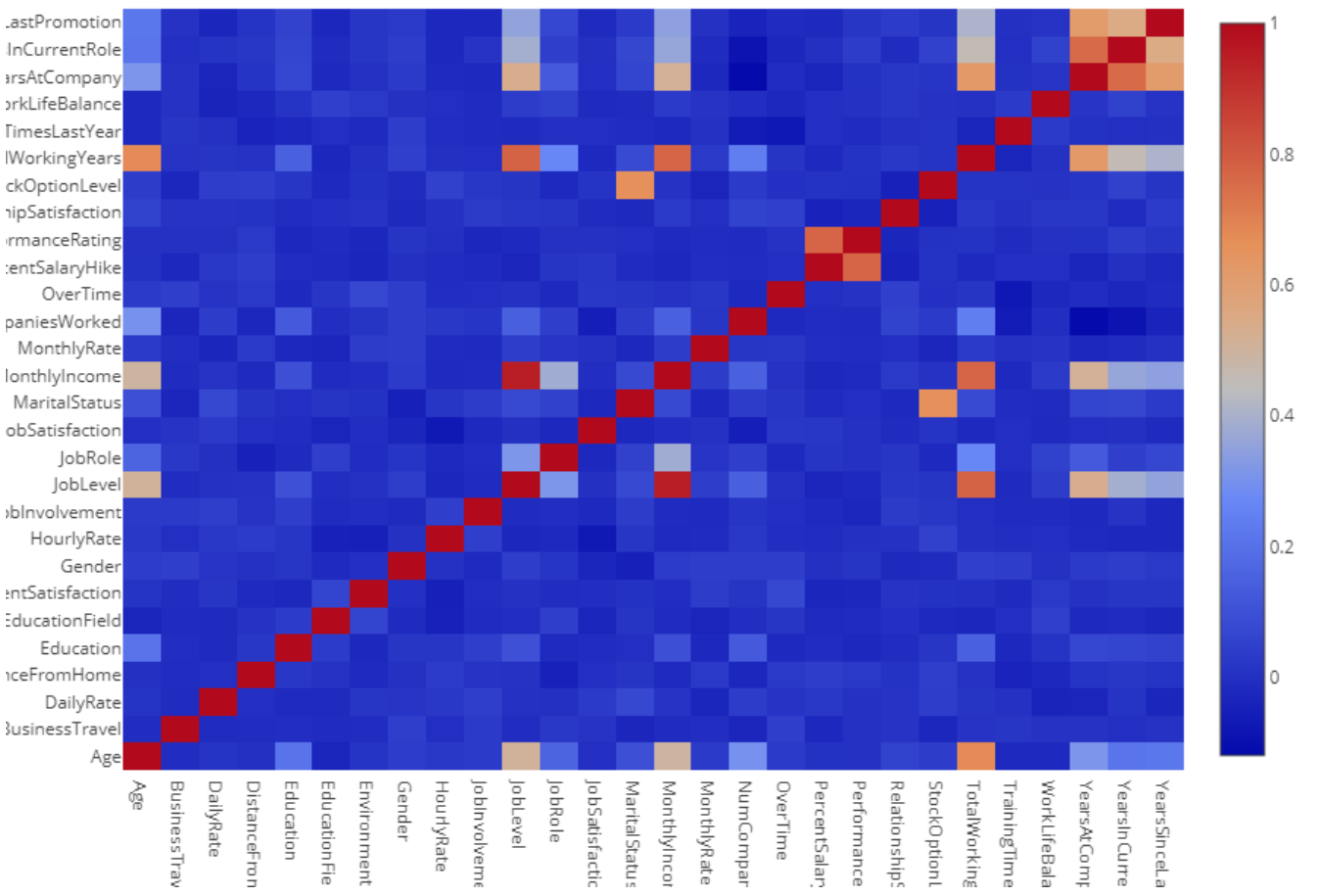

Fig. 1. Pearson Correlation of Numerical Values

As it is shown in fig. 1, higher correlated features are vivid from others with color map bar at the right. Henceforth, some of these features are compared together and analyzed to extract information. In addition to 2-Dimensional correlation analysis, we have visualized three features in a 3-Dimensioal space for better comprehension on changes of each axis.

## B. 2-Dimensianl Data Heatmap

### 1) Age Vs. JobLevel

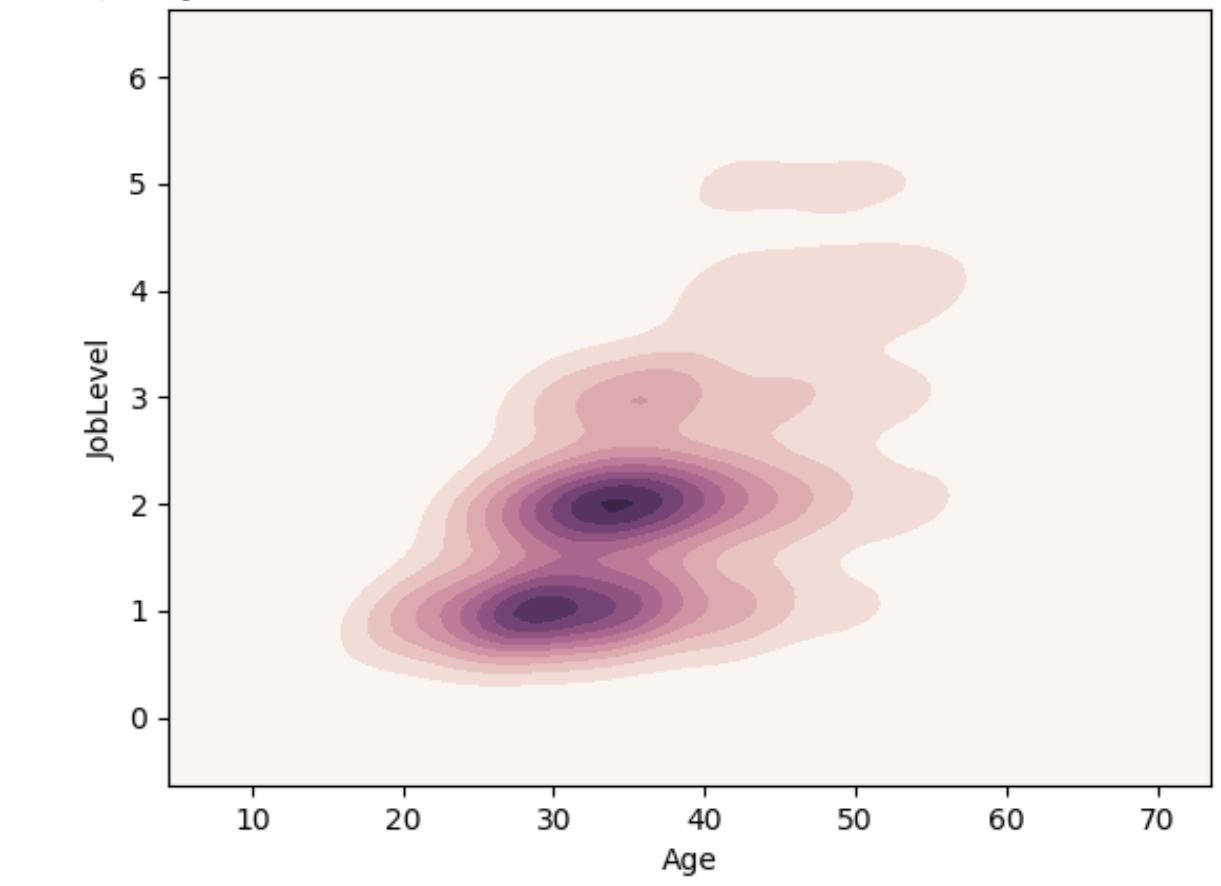

Fig. 2.

It is obvious from distribution of ages toward level of jobs that most of the employees are in rage 25 to 35 which have lower job roles.

### 2) JobLevel Vs. MonthlyIncome

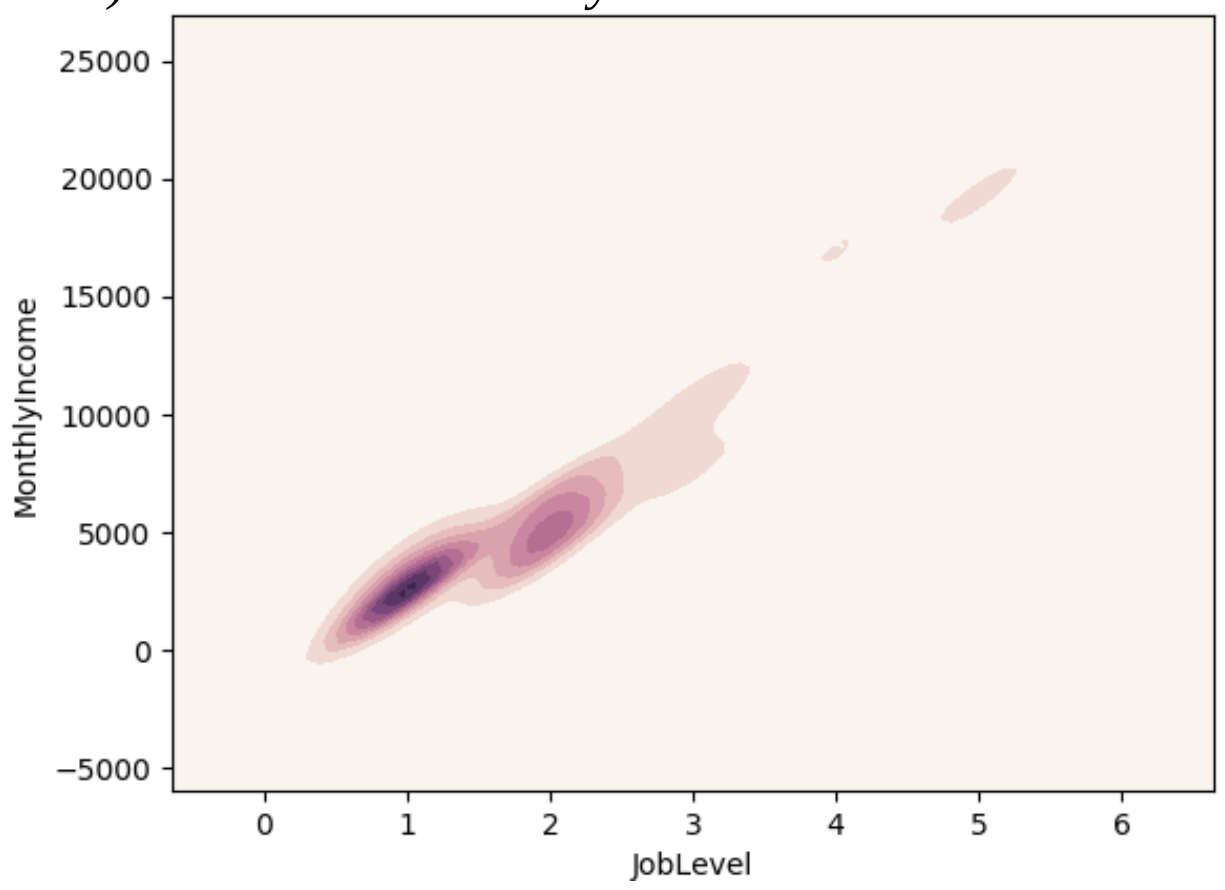

Fig. 3.

Most of the costs imposed to companies are related to employees with lower job levels. Although this cost is for monthly salary of employees. Cost of turnover of employees, which is more important, mostly happen in higher job levels with employees that have more experience. In spite of lower density in higher levels, they possess more value of human resources.

### 3) JobSatisfaction Vs. Attrition

In comparing employee satisfaction and turnover, it is apparent that increment in satisfaction of employees does not get along with zero attrition rate.

It is shown in fig. 5 that even higher satisfaction rates does not prevent employee to stay with their current job.

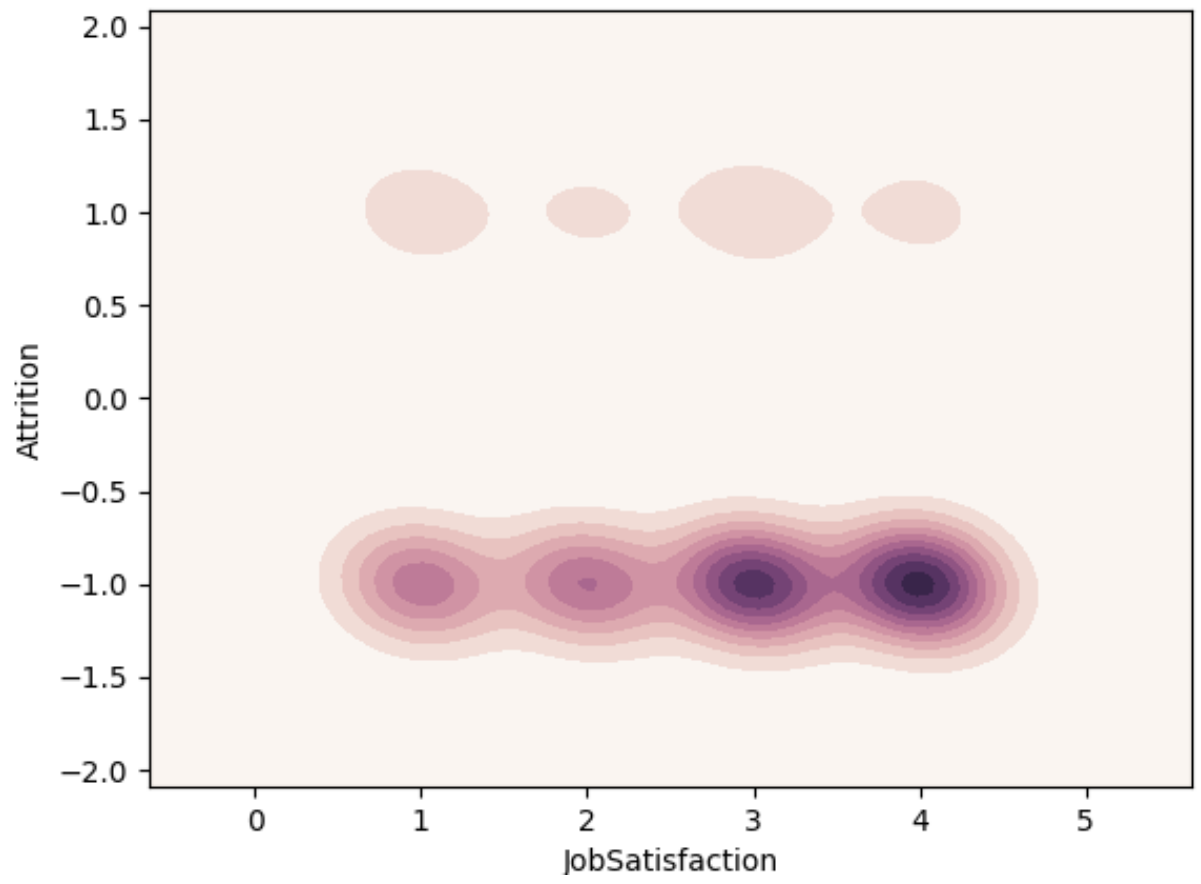

Fig. 4.

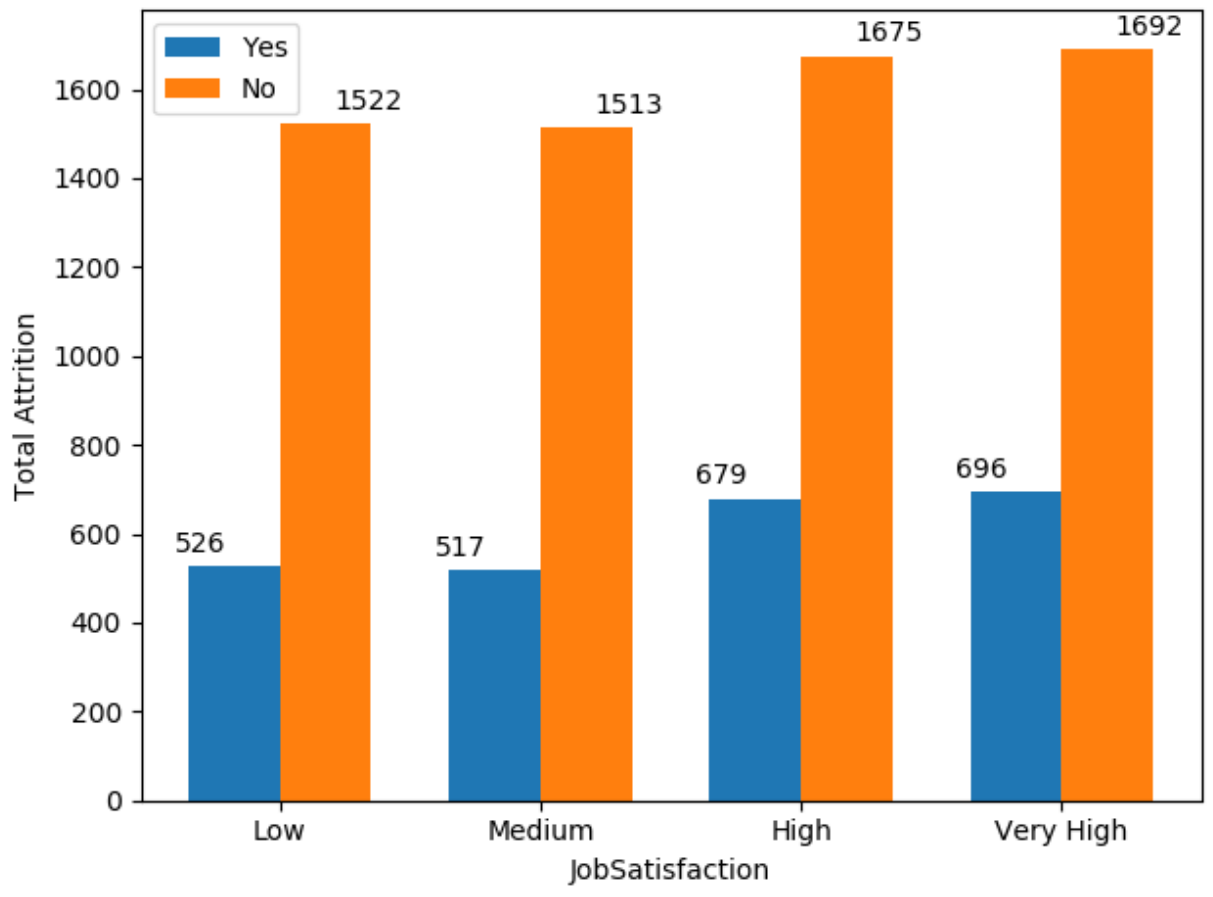

Fig. 5.

### 4) MaritalStatus Vs. StockOptionsLevel

Allocating stocks owned by an individual whom is single is almost zero. This can be used a job benefit factor for keeping different marital status types of employees satisfied with his/her current job profile.

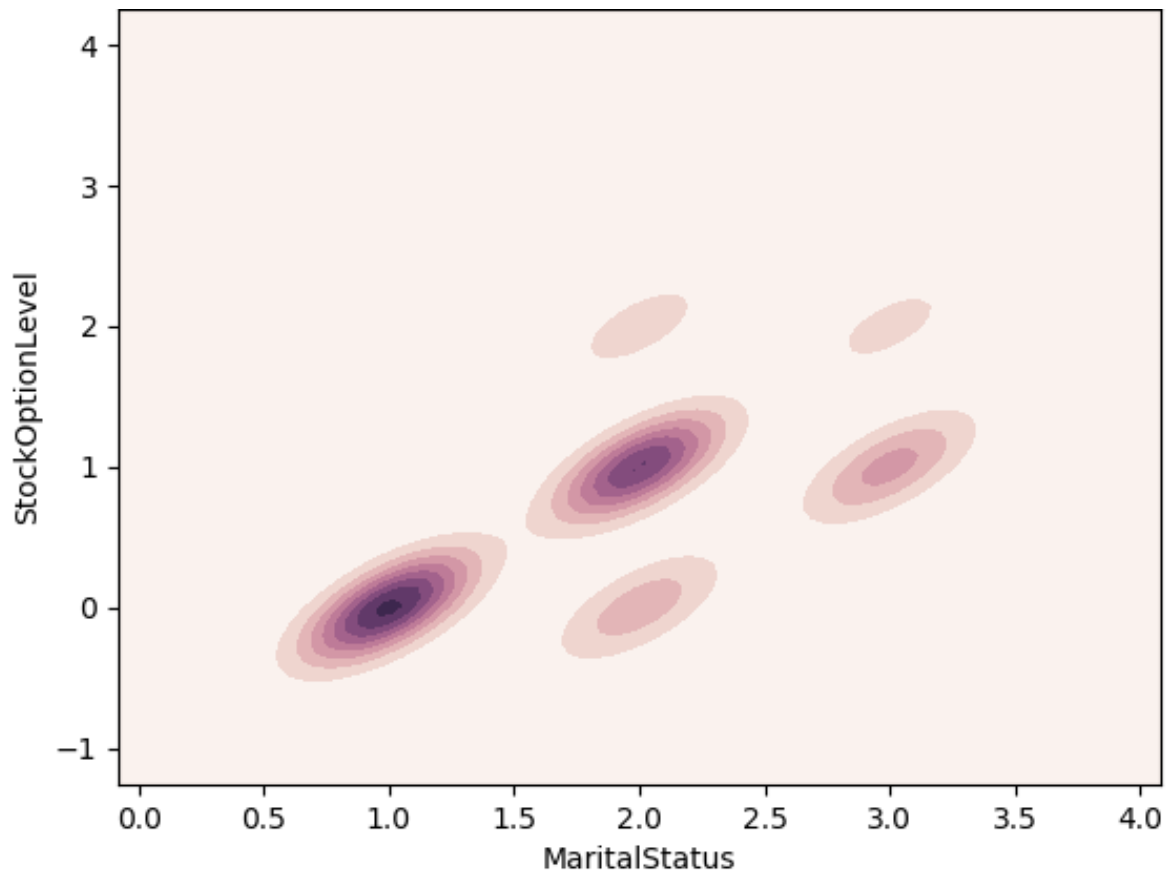

Fig. 6.

*5) MonthlyIncome Vs. Attrition*

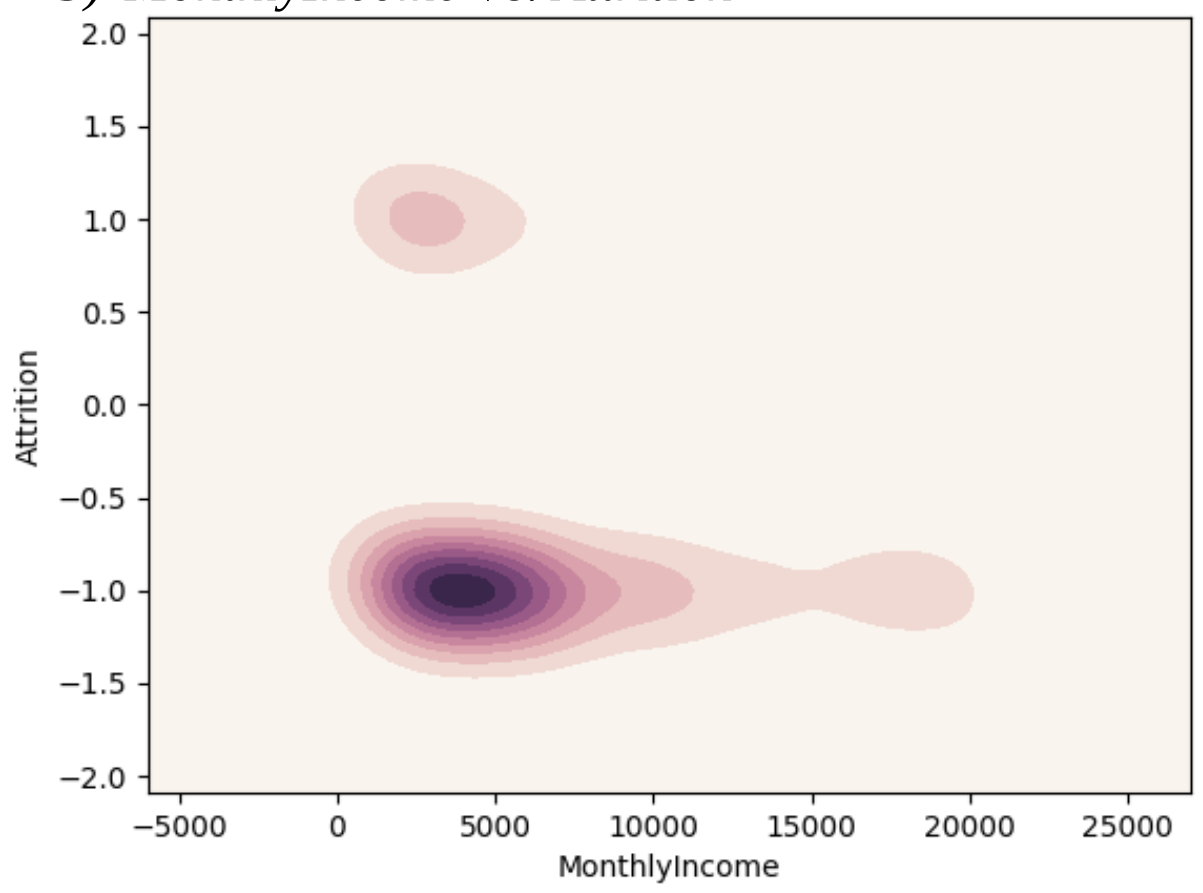

Fig. 7.

Most of the turnovers occur in lower salary rates. Higher job levels demand higher income for individuals. Keeping this asset in company is accomplished on behalf of growth in allocated human resources expenditures. This extra cost keeps the company knowledge-intense job roles with the company.

*6) PerformanceRating Vs. PercentSalaryHike*

Statistical analysis on percentage of increment on salary for employees with performance rating 3 (out of 4) is in range from 10 to 15 percent. While this number for employees with rate 4 is 10 percent higher in both lower and upper bounds.

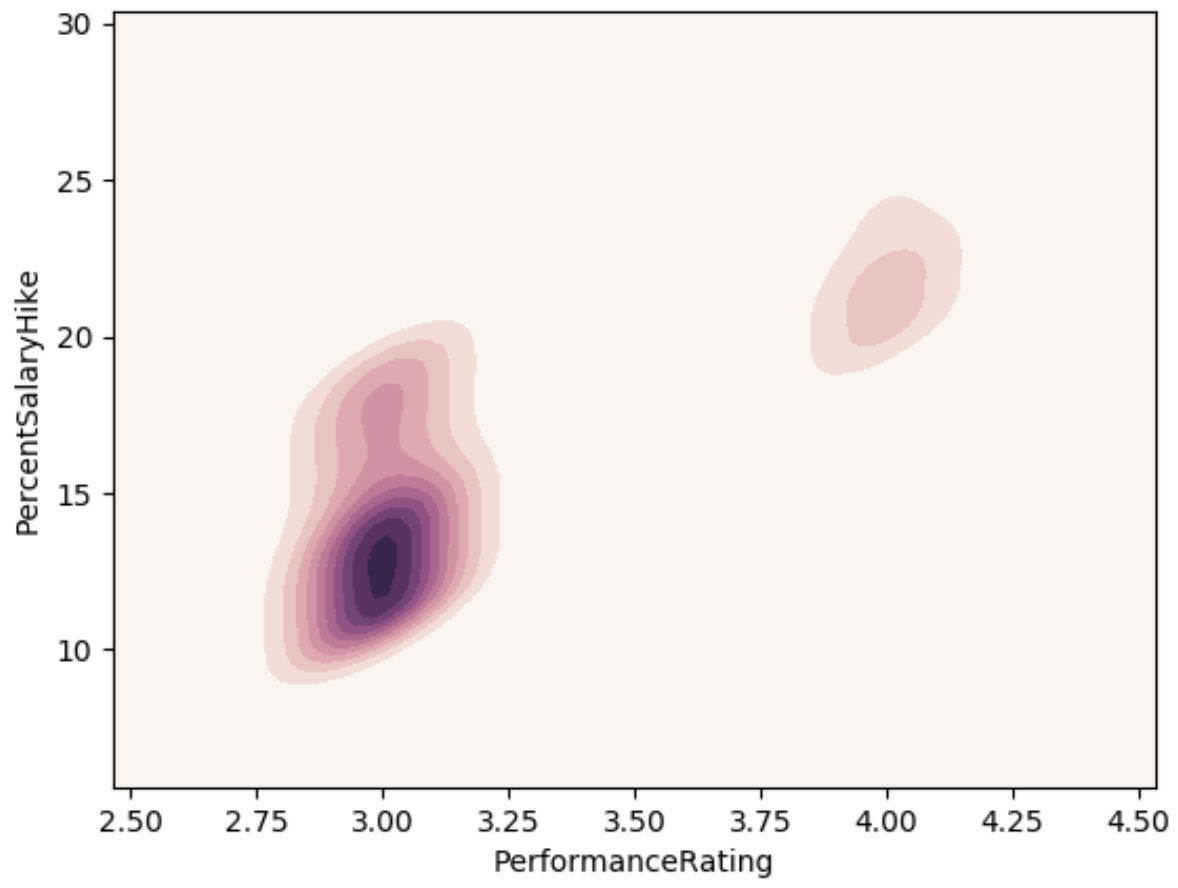

Fig. 8.

## C. 3-Dimensional Data Scatter

*1) MonthlyIncome Vs. TotalWorkingYear Vs. Attrition*

In comparing monthly income with total years an employee worked, distribution on higher salaries are getting less dense. but this is a 2 feature comparison, when it is compared with attrition, the amount of risk threatening company is focused on employees with approximately 10000 income and about 15 years of experience. Although this area is not as dense as lower incomes but the cost of losing resources with more than 10 years of experience is not agreeable.

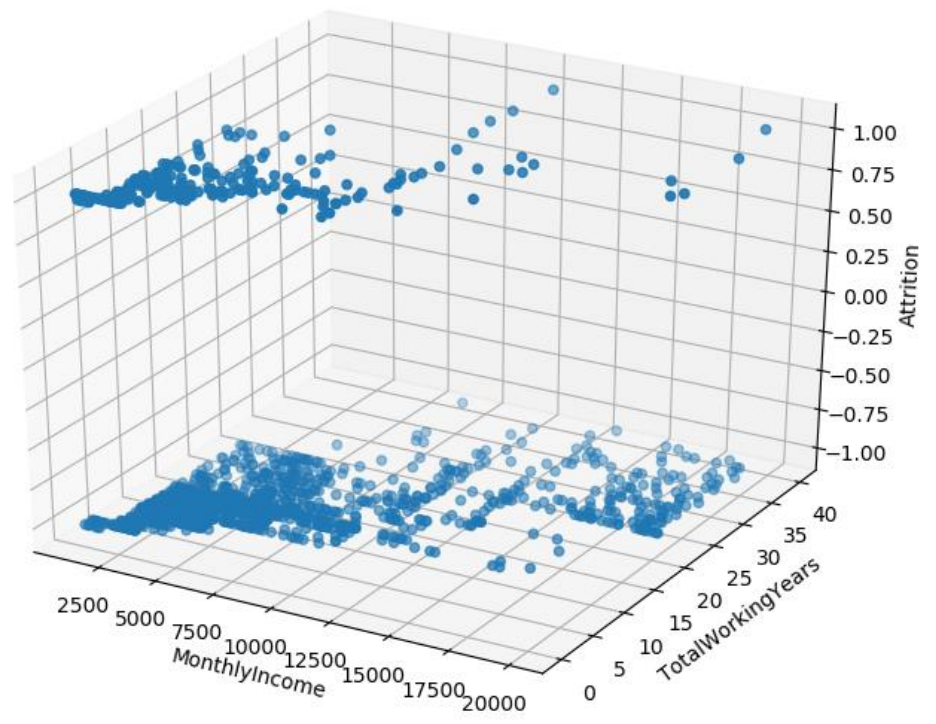

Fig. 9.

*2) TotalWorkingYears Vs. JobLevel Vs. Attrition*

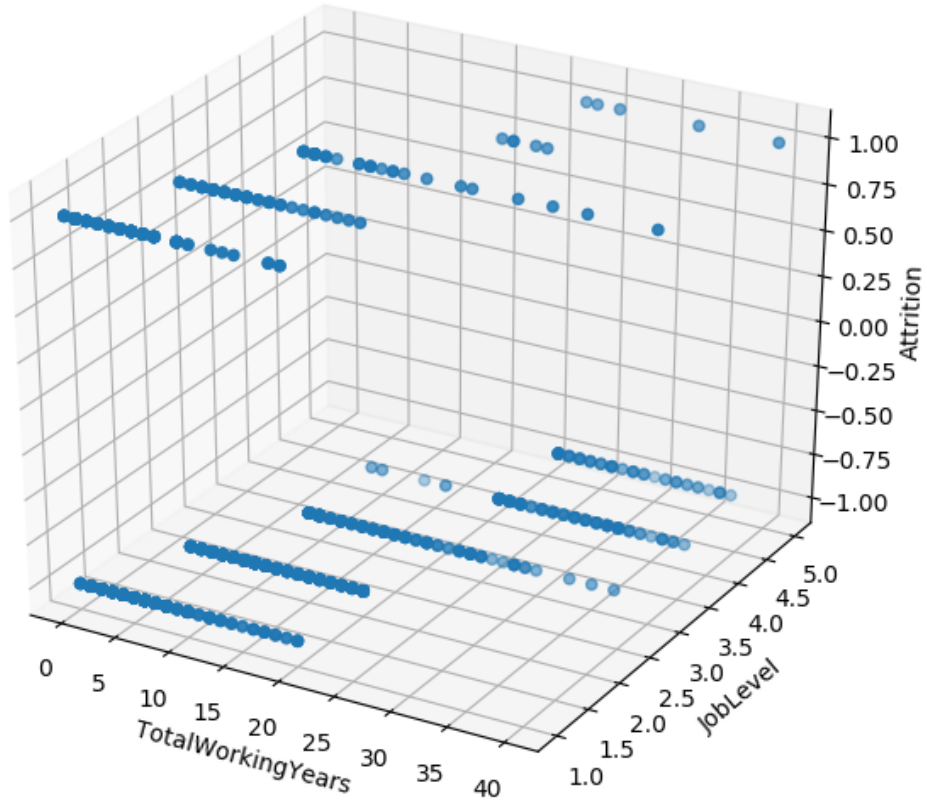

Fig. 10.

Fig. 10 specifies that higher job levels – which are more critical – require more working experience. like previous section, attrition in medium job levels with average of 15 years of experience can cause decrease in company's outcome.

## VII. Application of ML algorithms and results

In the following, the results of application of three classification algorithms are shown. Parameters of each algorithm are set to yield best accuracy. Optimal values are found with running algorithm multiple times.

AdaBoost has reached maximum accuracy among the others most of the time. With 1000 number of estimators and learning rate of 0.1, the accuracy obtained from multiple execution of algorithm on random splitting of training and test data was in range (0.84 0.88).

Second algorithm resulted accuracy in same range is Support Vector Machines. Both linear and radial basis function kernels resulted better than polynomial kernel. Penalty parameter C is set to 1 and gamma value for RBF kernel is $\frac{1}{Number\ of\ Instances}$. Based on the Occam's razor principle [16], and C. Cortes und V. Vapnik in 1995 [14], the linear kernel is preferred to RBF. To consider simplicity and avoid in increasing complexity of model which causes overfitting the results of SVM shown in table 2 are obtained with linear kernel rather than RBF.

RandomForest is the third algorithm used with 100 number of estimators. It also produces a model with same accuracy range.

| | ADABOOST | SVM | RANDOMFOREST |
|---|---|---|---|
| ITERATION 1 | 0.8707 | 0.8571 | 0.8503 |
| ITERATION 2 | 0.8775 | 0.8820 | 0.8798 |
| ITERATION 3 | 0.8820 | 0.8775 | 0.8775 |

TABLE II. ACCURACY SCORE OF THREE ITERATIONS EACH ON DIFFERENT TRAIN AND TEST SPLIT (RANDOMLY)

Afterwards, the Receiver Operating Characteristic of three classifiers in single plot expresses the performance of each model with given threshold. To better analyze this, we have shown iteration 1's ROC curve in fig 11.

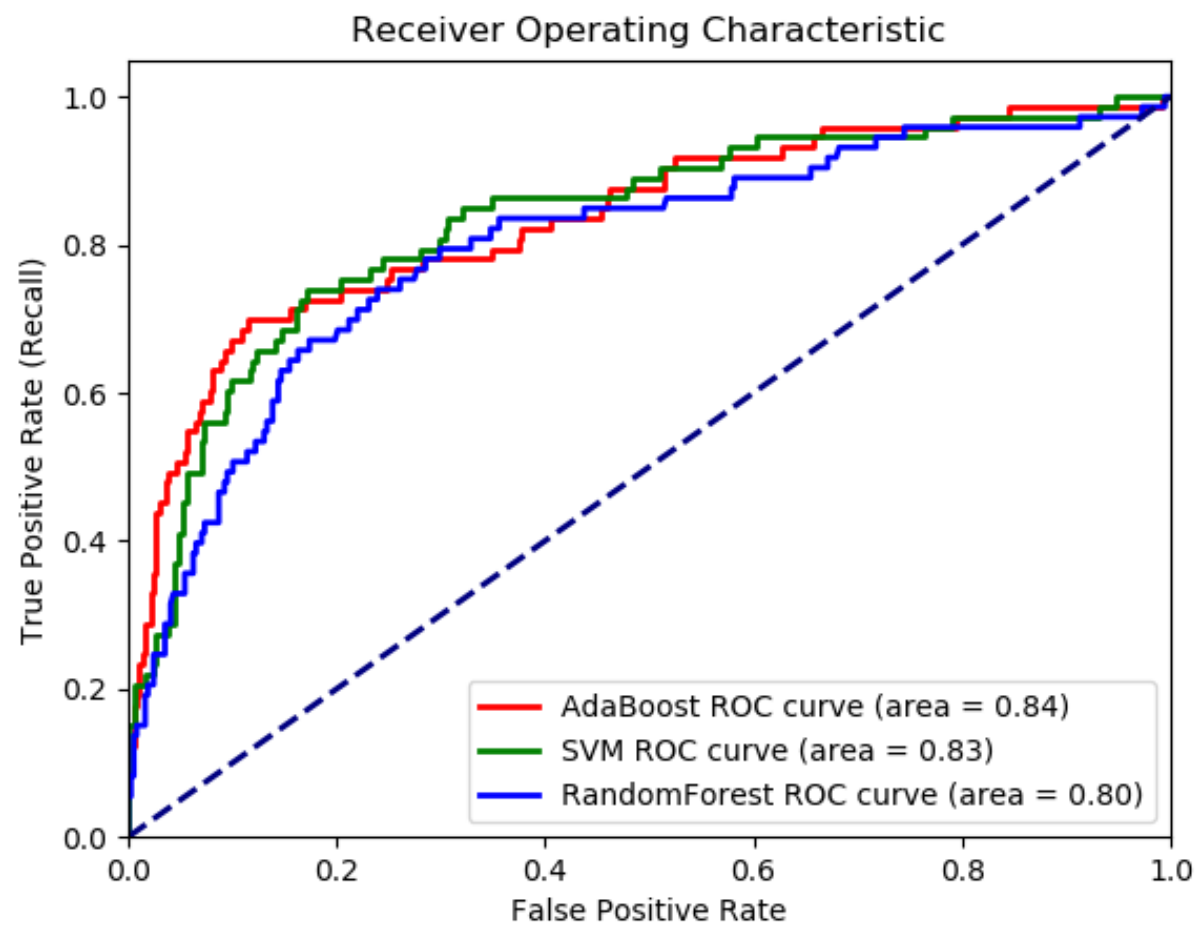

Fig. 11. ROC curve for iteration 1 of table 2.

The boosting algorithm AdaBoost has better performance until selection of TPR = 0.7 as threshold. The SVM classifier gain better performance as TPR gets higher. AdaBoost has highest area under curve shown in fig. 11. It means the chance for AdaBoost classifier to distinguish and label instances are 84% successful, which this probability for SVM and RandomForest are 83% and 80% respectively.

## VIII. FUTURE WORK

Many HR tasks can be automated, resulting in greater effectiveness with greater efficiency. For example, using AI, a company can take a job description, collect and analyze data from multiple sources, identify candidates who might never have even thought about applying for an open position, and contact them for an interview [17]. Tracking employee behaviors, such as computer use, is a way AI can be used to improve employee job performance and productivity. Many tasks require human-machine interactions such as voice-activated assistants and work place chatbots which is achieved by application of machine learning algorithms and speech recognition. AI can provide helpful information for decision making processes performed by individuals [18].

By using clustering algorithms (e.g., Fuzzy C-means), validation on classification labels of training data can be improved. This yields more realistic prediction.

## IX. CONCLUSIONS

One of the most important issue in Human Resource Management is employee attrition rate prediction. In this paper, we applied machine learning algorithms to employee's turnover. Pearson correlation between features has been illustrated. Highly correlated features compared in heatmap diagrams. Two ensemble learning algorithms and SVM used to create models. The results are populated in Table II comparing accuracy of training data between AdaBoost, SVM and RandomForest. Our models have reached maximum accuracy of %88. Fig. 11 illustrates three algorithms characteristics for different recall cut-offs.